# Attitude Reconstruction from Inertial Measurement: Mitigating Runge Effect for Dynamic Applications

Yuanxin Wu and Maoran Zhu

*Abstract*—Time-equispaced inertial measurements are practically used as inputs for motion determination. Polynomial interpolation is a common technique of recovering the gyroscope signal but is subject to a fundamentally numerical stability problem due to the Runge effect on equispaced samples. This paper reviews the theoretical results of Runge phenomenon in related areas and proposes a straightforward borrowing-and-cutting (BAC) strategy to depress it. It employs the neighboring samples for higher-order polynomial interpolation but only uses the middle polynomial segment in the actual time interval. The BAC strategy has been incorporated into attitude computation by functional iteration, leading to accuracy benefit of several orders of magnitude under the classical coning motion. It would potentially bring significant benefits to the inertial navigation computation under sustained dynamic motions.

## I. INTRODUCTION

Inertial sensors, i.e., gyroscopes and accelerometers, produce discrete measurements mostly at equally-spaced time instants, which are widely used to design algorithms for motion determination, e.g., in various applications of attitude estimation or inertial navigation [1, 2]. Usually, the inertial sensor signals are reconstructed by means of popular polynomial interpolation methods [3-11], as polynomials are supposed to approximate a smooth function as closely as possible if the number of samples are sufficiently large. The smooth interpolant of the underlying smooth functions at equispaced samples is a ubiquitous problem [12]. However, the close approximation of polynomials to a smooth function is preconditioned on the distribution of samples in addition to the function property [13]. It is well known that polynomial interpolants in Chebyshev samples, which are clustered towards the interval boundary, are well-behaved and have been widely used. For interpolants of a function at equispaced samples to converge along with the growing number of samples, the function must be analytic, not just on the considered interval but throughout the so-called Runge region [13]. Furthermore, computer rounding errors might be considerably amplified or even cause divergence in practice, depending on the distribution of the samples. Unfortunately, the equispaced samples are among those that are exponentially ill-conditioned in term of numerical stability [13]. The Runge phenomenon, in which wild oscillations are observed near the interval boundary, is not unusual for an equispaced high-order interpolant.

A number of methods have been proposed to approximate the underlying functions from equispaced samples, but unfortunately it was proven that even for analytic functions no stable method can be simultaneously exponentially convergent and well-conditioned [14]. The linear rational interpolant proposed in [15] has been well known as a significant development in the field of numerical analysis [12], and was further extended in [16] making use of additional samples beyond the interval boundary for much improved error mitigation in the original interval. Notably, it possesses error characteristics comparable to the excellent polynomial interpolant in Chebyshev samples and is very close to be optimal for equispaced samples [12].

Recent advance in the inertial navigation computation has been achieved by the functional iterative integration approach to solve the kinematics differential equations up to the machine precision [8-11], which would potentially bring significant benefits to the inertial navigation computation under sustained dynamic motions [17, 18]. As it takes as inputs the angular velocity/specific force polynomials fitted from time-equispaced gyroscope/accelerometer measurements, in principle any approximation error in the fitted polynomials would inevitably decay the computation accuracy. For instance, the Runge phenomenon is recently found having an obvious adverse effect on the accuracy and robustness of attitude computation for dynamic applications [19]. For a fixed sampling rate restricted by any practical inertial navigation system, more gyroscope samples might be desired for more accurate polynomial approximation of angular velocity, but it is fundamentally prohibited by the Runge phenomenon, as reported in [19].

The technical contribution of the current paper is proposing a straightforward strategy of suppressing the Runge effect in equispaced interpolation and incorporating it into attitude computation for enhanced accuracy in highly dynamic applications. It helps break the barrier of high-accuracy senor measurement interpolation on equispaced samples for rigid motion reconstruction. The content is organized as follows. Section II briefly reviews the polynomial and rational interpolation advances in the applied mathematics community, and specifically underlines their numerical properties in terms of Lebesgue constant by a simple example. Section III investigates polynomial and rational interpolations in angular velocity reconstruction and attitude computation. Conclusions are drawn in the last section.

The work was supported in part by National Key R&D Program of China (2018YFB1305103). An abstract was submitted to DGON ISS 2021, Germany.

Authors' address: Shanghai Key Laboratory of Navigation and Location-based Services, School of Electronic Information and Electrical Engineering, Shanghai Jiao Tong University, Shanghai, China, 200240, E-mail: (yuanx_wu@hotmail.com).

## II. Polynomial/Rational Interpolation and Runge Effect

For the ordered distinct nodes $a = t_0 < t_1 < \ldots < t_N = b$ and the corresponding signal values $f_0, f_1, \ldots, f_N$, this section reviews the technique of polynomial and rational interpolations for obtaining functional reconstruction of the original signal. Specifically, the underlying reason of the Runge effect is highlighted.

### A. Polynomial Interpolation

For the above $N+1$ nodes associated with signal values, there is a unique interpolation polynomial $p_N$ of degree at most $N$, which interpolates $f$ satisfying $p_N(t_j) = f_j$ for $0 \le j \le N$. There are several ways to construct and represent the polynomial, for example, in the Lagrange form as [20]

$$p_N(t) = \sum_{j=0}^{N} \ell_j(t) f_j, \quad \ell_j(t) = \prod_{\substack{k=0 \\ k \ne j}}^{N} \frac{(t - t_k)}{(t_j - t_k)} \quad (1)$$

It can be readily checked that $\ell_j(t_i) = 1$ when $i = j$ and $\ell_j(t_i) = 0$ otherwise. Denote $\ell(t) = \prod_{k=0}^{N}(t - t_k)$. For $t$ other than the above distinct nodes, the polynomial (1) can be reformulated as

$$p_N(t) = \ell(t) \sum_{j=0}^{N} \frac{w_j}{(t - t_j)} f_j, \quad w_j = \prod_{\substack{k=0 \\ k \ne j}}^{N} \frac{1}{(t_j - t_k)} = 1/\ell'(t_j) \quad (2)$$

For the constant signal $f(t) \equiv 1$, we have

$$1 \equiv p_N(t) = \ell(t) \sum_{j=0}^{N} \frac{w_j}{(t - t_j)} \quad (3)$$

from the uniqueness of the polynomial interpolation. Dividing (2) by (3) gives the well-known barycentric form of the polynomial interpolant [12]

$$p_N(t) = \sum_{j=0}^{N} \frac{w_j}{(t - t_j)} f_j \Big/ \sum_{j=0}^{N} \frac{w_j}{(t - t_j)} \quad (4)$$

with the special case $p_N(t) = f_j$ for $t = t_j$. Though with the denominator $t - t_j$, it can be shown that $\lim_{t \to t_j} p_N(t) = f_j$ for $0 \le j \le N$, which means (4) is provably numerically stable [12, 13]. Seemingly like a rational function, (4) is indeed a polynomial with the specially defined weights given as in (2). For other weights, however, it will in general be a rational function instead of a polynomial, which is to be discussed in the sequel. The above barycentric form has the advantage of cancelling common factors and resulting in simple weights [13]. Regarding the equispaced nodes, the weights are $w_j = (-1)^j \binom{N}{j}$; for the Chebyshev nodes of the second kind, namely, $t_j = \cos(j\pi/N)$ for $0 \le j \le N$, the weights are $w_j = (-1)^j$ for $1 \le j \le N-1$ and $w_j = (-1)^j / 2$ for

$j = 0, N$.

The error of an interpolant is governed by not only the theoretical interpolation error in exact arithmetic but the amplification of rounding/measurement error [21]. The latter is characterized by the Lebesgue constant

$$\Lambda_N = \max_{a \le t \le b} \Lambda_N(t) = \max_{a \le t \le b} \sum_{j=0}^{N} \left| \frac{w_j}{(t - t_j)} \right| \Big/ \left| \sum_{j=0}^{N} \frac{w_j}{(t - t_j)} \right| \quad (5)$$

which is defined as the maximum of the Lebesgue function $\Lambda_N(t)$. Suppose the signal values $f_j$ are contaminated by a relative perturbation $\varepsilon_j f_j$, where $|\varepsilon_j|$ are not greater than some positive $\varepsilon$. Then the interpolant based on the perturbed values satisfies [21]

$$\begin{aligned}
\|f - \tilde{p}_N\|_\infty &= \max_{a \le t \le b} \left| f(t) - \sum_{j=0}^{N} \frac{w_j}{(t - t_j)}(f_j + \varepsilon_j f_j) \Big/ \sum_{j=0}^{N} \frac{w_j}{(t - t_j)} \right| \\
&\le \|f - p_N\|_\infty + \varepsilon \max_{a \le t \le b} \sum_{j=0}^{N} \left| \frac{w_j f_j}{(t - t_j)} \right| \Big/ \left| \sum_{j=0}^{N} \frac{w_j}{(t - t_j)} \right| \\
&\le \|f - p_N\|_\infty + \varepsilon \|f\|_\infty \Lambda_N
\end{aligned}$$
$$(6)$$

in which the first term is related to the theoretical interpolation error in exact arithmetic and the second term is related to the interpolation error in presence of rounding/measurement error. It has been proved that the Lebesgue constant $\Lambda_N > 2^{N-2}/N^2$ for equispaced nodes and $\Lambda_N \le 2\log(N+1)/\pi + 1$ for Chebyshev nodes [13]. No interpolation method for equispaced nodes exists that converges geometrically and has a Lebesgue constant decreasing geometrically [14, 21]. As a matter of fact, the polynomial interpolation in equispaced points is exponentially ill-conditioned, which leads to the well-known Runge phenomenon near the interval boundary. See a demo example (polynomial interpolation in red lines in Fig. 2) illustrated in Sec. II.C below.

### B. Rational Interpolation

The rational interpolation methods can be classified in two classes: nonlinear ones and linear ones [12]. In the latter case, the interpolant is linear in $f$, resembling the form in (4) of which the denominator depends on the nodes but not on $f$. Notably, the famous family of Floater and Hormann (FH) rational interpolant [15], for a design parameter $d$, is given by

$$r_N(t) = \sum_{j=0}^{N-d} \lambda_j(t) p_{d[j]}(t) \Big/ \sum_{j=0}^{N-d} \lambda_j(t), \quad \lambda_j(t) = \frac{(-1)^j}{(t - t_j) \cdots (t - t_{j+d})} \quad (7)$$

which blends $N - d + 1$ polynomials $p_{d[j]}$ with $\lambda_j$ acting as the weighting functions. Note that $p_{d[j]}$ interpolates $d+1$ consecutive nodes starting from $t_j$, in other words, it is the polynomial of degree at most $d$ interpolating $f_j, f_{j+1}, \ldots, f_{j+d}$. It can be seen that (7) reduces to the polynomial interpolation (1) for $d = N$. The FH rational interpolant can be rewritten as the barycentric form [15]

$$r_N(t) = \sum_{j=0}^{N}\frac{w_j}{\left(t-t_j\right)}f_j \Big/ \sum_{j=0}^{N}\frac{w_j}{\left(t-t_j\right)} \tag{8}$$

with weights

$$w_j = \sum_{i\in J_j}(-1)^i\prod_{\substack{k=i\\k\neq j}}^{i+d}\frac{1}{t_j-t_k}, \quad J_j = \left\{i\in\{0,\ldots,N-d\}: j-d\leq i\leq j\right\} \tag{9}$$

Table I. Lebesgue constants of polynomial and rational interpolation

| Interpolation on Different Nodes | Polynomial Interpolation | | Rational Interpolation | |
|---|---|---|---|---|
| | *Equispaced* | *Chebyshev* | *Equispaced (FH)* | *Equispaced (EFH)* |
| Lebesgue Constant | $\Lambda_N > 2^{N-2}/N^2$ | $\Lambda_N \leq 2\log(N+1)/\pi + 1$ | $\dfrac{2^{d-2}}{d+1}\ln\left(\dfrac{N}{d}-1\right) \leq \Lambda_N \leq 2^{d-1}\left(2+\ln N\right)$ | $\Lambda_N \leq 2 + \ln(N+2d)$ |

It reduces to $w_j = (-1)^{j-d}\sum_{i\in J_j}\binom{d}{j-i}$ for equispaced nodes.

As long as $f$ is smooth enough, the FH interpolant has the convergence rate $O\left(h^{d+1}\right)$ independent of how the nodes are distributed, where $h$ is the maximum spacing between consecutive nodes. Larger value of $d$ is supposed to lead to faster convergence, but rounding or measurement errors also come into effect. For equispaced nodes, it was shown in [22] that the Lebesgue constant for the FH interpolant (8)-(9) is $O\left(2^d\ln N\right)$, which grows logarithmically with $N$ and exponentially with $d$.

An interesting feature was found in [16] that the Lebesgue function $\Lambda_N(t)$ of the FH interpolant has at most $d$ oscillations at both ends and is much smaller in the remaining part of the interval. Consequently, additional $d$ signal values are generated by a numerical Taylor expansion on both ends of the interval; the whole new set of signal values are interpolated by a rational function

$$r_{N+2d}^*(t) = \sum_{j=-d}^{N+d}\frac{w_j^*}{\left(t-t_j^*\right)}f_j^* \Big/ \sum_{j=-d}^{N+d}\frac{w_j^*}{\left(t-t_j^*\right)} \tag{10}$$

but only evaluated in the original interval $[a,b]$. The superscript * is used to denote the extended family of FH (EFH) interpolant. The weights are computed by means of (9) yet for $N+2d+1$ nodes. Notably, the associated Lebesgue constant for the extended FH interpolant is $O\left(\ln(N+2d)\right)$ for equispaced nodes, i.e., growing logarithmically with $N$ and $d$. Remarkable improvement of interpolation accuracy over the original FH interpolant was demonstrated in [16]. It was conjectured that the extended FH interpolant is likely very close to optimality for equispaced nodes [12].

The strategy of using the neighbouring samples to assist signal reconstruction is not uncommon in the navigation community. For example, actual gyroscope samples in the previous computation interval, in contrast to the Taylor-expansion extrapolated points in EFH [16], have been frequently used in designing attitude algorithms [23-25]. For the attitude computation of main interest in this paper, the required additional signal values on both ends of the current interval would be actual gyroscope samples at the cost of moderate computational latency. To use the neighbouring samples is specifically named as the borrowing-and-cutting (BAC) strategy for interpolation. This name is used in this paper to intuitively underline the main idea of the approach.

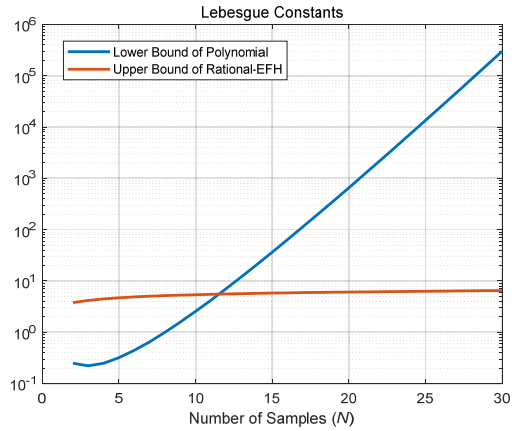

Figure 1. Bounds of Lebesgue constants for polynomial and EFH rational ($d = N$) interpolations on equispaced nodes.

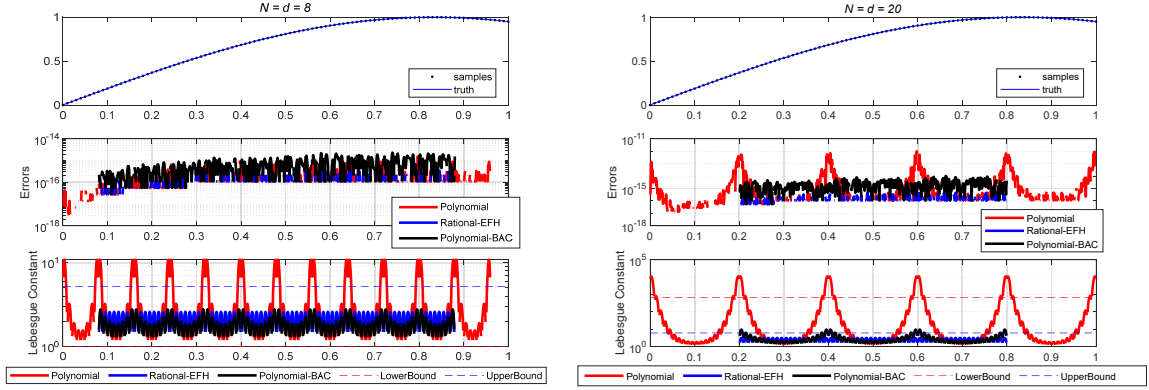

Figure 2. Signal $\sin(0.6\pi t)$ reconstruction errors and Lebesgue functions of polynomial and rational (EFH) interpolations for $N = d = 8$ and $N = d = 20$. Polynomial-BAC indicates polynomial interpolation using the additional samples beyond current time interval.

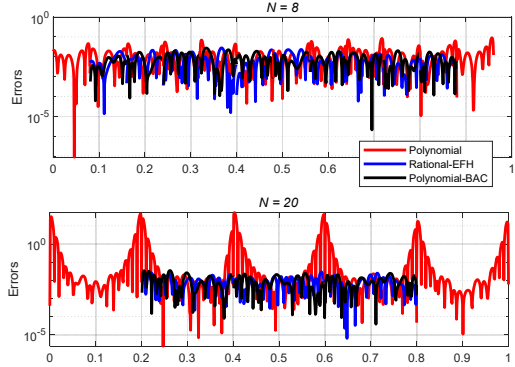

Figure 3. Signal $\sin(0.6\pi t)$ reconstruction errors by polynomial and rational (EFH) functions of noise-contaminated samples for $N = d = 8$ and $N = d = 20$. Polynomial-BAC indicates polynomial interpolation using additional samples beyond current time interval.

## C. Polynomial/Rational Reconstruction Example on Equispaced Nodes

Table I lists the Lebesgue constant bounds for polynomial/rational interpolation on equispaced and Chebyshev nodes. The Lebesgue constant bounds at equispaced samples are plotted in Fig. 1. The bound trend tells that the Lebesgue constant of the polynomial interpolation is by far larger than that of the EFH rational interpolation when the number of samples surpasses ten. Note that the lower bound for the polynomial seems quite overoptimistic, as will be seen from Fig. 2 below. Generally speaking, it is safe for us to use the polynomial interpolation for $N \leq 10$. This observation is consistent with the selection of eight samples in attitude/navigation computation in [8-11].

Next, we examine the polynomial and EFH rational interpolation reconstruction quality for a trigonometric signal $f(t) = \sin(0.6\pi t)$. The equispaced sampling is performed at 100 Hz over the time interval $\begin{bmatrix} 0 & 1 \end{bmatrix}$. Figure 2 presents the polynomial and rational interpolation results for $N = d = 8$ and $N = d = 20$, as well as their Lebesgue functions $\Lambda_N(t)$ and corresponding upper/lower bounds as given in Table I. Note that the EFH rational interpolation method requires $N + 2d + 1$ samples to complete interpolation (with extra $d$ samples on both sides), so there exist interpolation gaps on interval boundaries in Fig. 2. The EFH rational interpolation produces comparable errors for both $N = d = 8$ and $N = d = 20$, while the polynomial interpolation for $N = d = 20$ leads to much larger errors at interval boundaries by about four orders of magnitude. The Lebesgue constant of the EFH rational is consistently small, but that of the polynomial significantly deteriorates up to about three orders from $N = d = 8$ to $N = d = 20$. It appears that the polynomial interpolation also has the same feature as reported in [16], namely, its errors or Lebesgue constants are big near both ends but much smaller in the middle part. Therefore, we apply the EFH's BAC strategy to the polynomial interpolation and plot the result in Fig. 2 as well. Normal polynomial, polynomial-BAC and EFH rational interpolations are comparable in interpolation accuracy for $N = d = 8$, while the EFH rational and polynomial-BAC interpolations become significantly better than the normal polynomial for $N=d=20$. Compared with the EFH rational interpolation, the polynomial-BAC interpolation has comparable Lebesgue function across the time interval and is slightly worse in accuracy by no more than one order of magnitude.

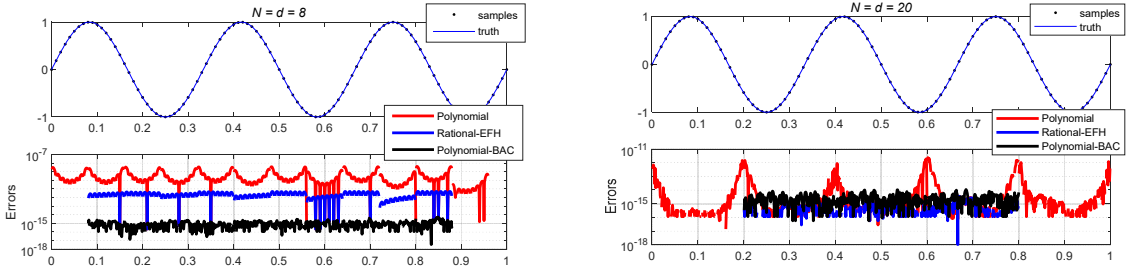

Figure 4. Signal $\sin(6\pi t)$ reconstruction errors of polynomial and rational (EFH) interpolations for $N=d=8$ and $N=d=20$. Polynomial-BAC indicates polynomial interpolation using additional samples beyond current time interval.

Figure 3 further assesses the interpolation performance with samples error-contaminated by a Gaussian noise of standard deviation 0.01. It can be seen from Figs. 2-3 that the Lebesgue constants well characterize the error behaviors of the polynomial and EFH rational interpolations, and the accuracy advantage of EFH rational interpolation for $N=d=20$ is still prominent in this error-contaminated scenario, so is the polynomial-BAC interpolation. In other words, the signal reconstruction quality at the presence of rounding/measurement errors is mostly governed by the Lebesgue constant as well. The advantage of EFH rational and polynomial-BAC interpolations would not be overwhelmed by the rounding/measurement errors.

Figure 4 plots the results for $f(t)=\sin(6\pi t)$ of frequency ten times larger. The results of $N=d=20$ change little as compared with the counterpart in Fig. 2, while a great discrepancy is observed for that of $N=d=8$. The normal polynomial seems unable to cope with the signal of higher frequency by using only $N+1=9$ samples, i.e., the polynomial order is 8, while the polynomial-BAC interpolation reaches a high order of 24 by using $N+2d+1=25$ samples. As the EFH rational interpolant is a blend of polynomials of order 8, the insufficient representation of the normal polynomial is also shared by the EFH rational interpolant. Note the overlapped spikes between polynomial and EFH rational interpolations.

## III. APPLICATION TO ANGULAR VELOCITY RECONSTRUCTION AND ATTITUDE COMPUTATION

This section applies the EFH rational and polynomial-BAC interpolations to the functional iterative attitude computation [8-10, 19]. Gyroscopes have two common output forms: angular velocity measurement and angular increment (integrated angular velocity) measurement. This paper takes the latter case as a demonstration.

### A. Angular Velocity Reconstruction

Assume discrete angular increment measurements $\Delta\tilde{\boldsymbol{\theta}}_{t_j}$ by a triad of gyroscopes, at equispaced time instants $t_j$ ($j=0,1,2,\ldots,N$) in the current time interval $\begin{bmatrix} 0 & t_N \end{bmatrix}$. With additional discrete angular increment measurements on both sides ($j=-d,\ldots,-1$ for the previous time interval and $j=N+1,\ldots,N+d$ for the subsequent time interval), we can reconstruct the integrated angular velocity function by the BAC-based polynomial or rational interpolation of the barycentric form (10)

$$\boldsymbol{\theta}^*_{N+2d}(t)=\sum_{j=-d}^{N+d}\frac{w_j^*}{\left(t-t_j^*\right)}\tilde{\boldsymbol{\theta}}^*_{t_j}\Bigg/\sum_{j=-d}^{N+d}\frac{w_j^*}{\left(t-t_j^*\right)}, \quad t\in\begin{bmatrix} 0 & t_N \end{bmatrix}$$
(11)

where $\tilde{\boldsymbol{\theta}}^*_{t_j}$ denotes the accumulated angular increments up to time instant $t_j$, i.e., $\tilde{\boldsymbol{\theta}}^*_{t_j}=\sum_{k=-d}^{j}\Delta\tilde{\boldsymbol{\theta}}_{t_j}$. For the polynomial-BAC interpolation, $w_j^*=(-1)^j\binom{N+2d}{j}$, while for the EFH rational interpolation, $w_j^*=(-1)^{j-d}\sum_{i\in J_j}\binom{d}{j-i}$ where $J_j=\left\{i\in\{-d,\ldots,N\}:j-d\le i\le j\right\}$.

As the reconstructed integrated angular velocity are normal polynomial or rational functions, we need to map the current time interval $\begin{bmatrix} 0 & t_N \end{bmatrix}$ onto $\begin{bmatrix} -1 & 1 \end{bmatrix}$ by letting $t=t_N(1+\tau)/2$ and then approximate the reconstructed function via Chebyshev polynomials, as required by attitude computation with the functional iteration approach [8-11]. The Chebyshev polynomial approximations of the reconstructed integrated angular velocity can be readily given as

$$\boldsymbol{\theta}^*_{N+2d}(t)\overset{t\to t_N(1+\tau)/2}{=}\boldsymbol{\theta}^*_{N+2d}(\tau)\approx\sum_{i=0}^{n_\theta}\mathbf{h}_i F_i(\tau)$$
(12)

where $F_i(x)$ is the $i$-th-degree Chebyshev polynomial of the first kind, and $n_\theta$ is the maximum order of Chebyshev polynomials [20]. As $N+2d+1$ samples are used for interpolation, it is enough for $n_\theta$ to be no more than $N+2d$. The Chebyshev coefficients in (12) can be obtained by [20]

$$\mathbf{h}_i \approx \frac{2-\delta_{0i}}{P_\theta} \sum_{k=0}^{P_\theta-1} \cos\frac{i(k+1/2)\pi}{P_\theta} \boldsymbol{\theta}_{N+2d}^* \left(\cos\frac{(k+1/2)\pi}{P_\theta}\right)$$ (13)

where $\delta_{0i}$ is the Kronecker delta function, yielding 1 for $i=0$ and zero otherwise. Exact coefficients could be obtained only if the number of summation terms, $P_\theta$, approaches infinity. The angular velocity estimate is to be acquired by time derivative of the integrated angular velocity function in (12), with the aid of the derivative property of the Chebyshev polynomial of the first kind [26]

$$\hat{\boldsymbol{\omega}}(\tau) \triangleq \left[\frac{d}{dt}\boldsymbol{\theta}_{N+2d}^*(t)\right]\Bigg|_{t \to t_N(1+\tau)/2} = \frac{2}{t_N}\frac{d}{d\tau}\boldsymbol{\theta}_{N+2d}^*(\tau)$$

$$\approx \frac{2}{t_N}\sum_{i=0}^{n_\theta}\mathbf{h}_i\frac{d}{d\tau}F_i(\tau) \triangleq \frac{2}{t_N}\sum_{i=0}^{n_\theta}i\mathbf{h}_iU_{i-1}(\tau)$$ (14)

where

$$U_j(\tau) = \begin{cases} 2\sum_{\text{odd } s}^{j}F_s(\tau) & \text{for odd } j \\ 2\sum_{\text{even } s}^{j}F_s(\tau)-1 & \text{for even } j \end{cases}$$ (15)

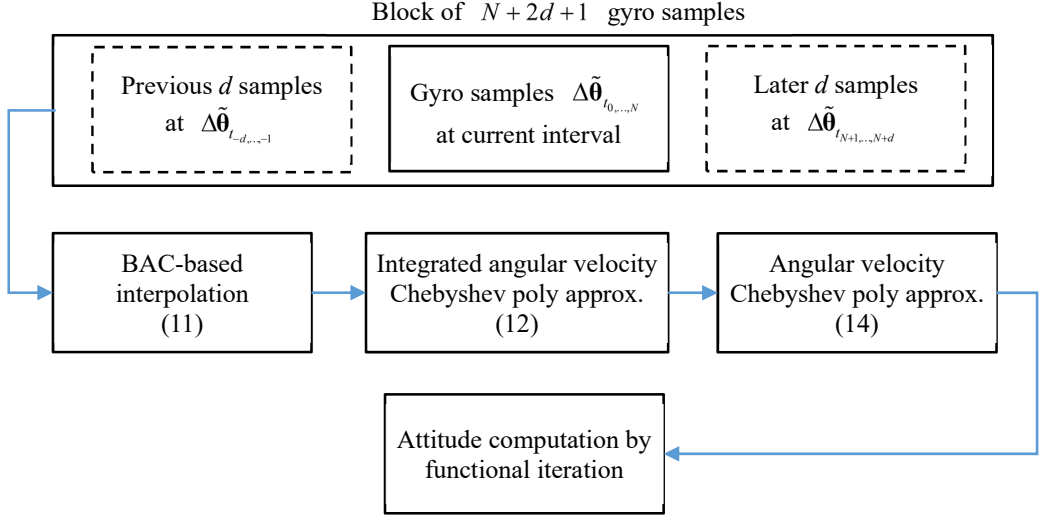

Figure 5. Flowchart of BAC-based interpolation for angular velocity reconstruction and attitude computation.

Figure 5 illustrates the flowchart of the angular velocity polynomial/rational reconstruction and the Chebyshev polynomial approximation, which is to be used as the input of the subsequent attitude computation by functional iteration.

### B. Numerical Results of Attitude Computation

The classical coning motion, widely accepted as the standard algorithm criterion, is used here to evaluate the interpolation quality and its effect on attitude computational accuracy. The angular velocity of the classical coning motion is described by $\boldsymbol{\omega} = \Omega\begin{bmatrix} -2\sin^2(\alpha/2) & -\sin(\alpha)\sin(\Omega t) & \sin(\alpha)\cos(\Omega t) \end{bmatrix}^T$, with the true rotation vector $\boldsymbol{\sigma} = \alpha\begin{bmatrix} 0 & \cos(\Omega t) & \sin(\Omega t) \end{bmatrix}^T$ and the true attitude quaternion $\mathbf{q} = \cos(\alpha/2) + \sin(\alpha/2)\begin{bmatrix} 0 & \cos(\Omega t) & \sin(\Omega t) \end{bmatrix}^T$. In the above, $\alpha$ denotes the coning angle and $\Omega = 2\pi f_c$ denotes the angular frequency and $f_c$ is the coning frequency. The angular increment measurement is assumed and the sampling rate is nominally set to $f_s = 1000$ Hz. The following principal angle metric is used to quantify the attitude computation error

$$\varepsilon_{att} = 2\left\|\left[\mathbf{q}^* \circ \hat{\mathbf{q}}\right]_{2:4}\right\|$$ (16)

where $\hat{\mathbf{q}}$ denotes the quaternion estimate computed by attitude algorithms, $\circ$ means the quaternion multiplication, and the operator $[\cdot]_{2:4}$ takes the vector part of the error quaternion. The attitude computation by functional iterative approach is exemplified by QuatFIter [10] hereafter.

Figure 6 compares the attitude errors of polynomial, EFH rational and polynomial-BAC angular velocity reconstruction for the first ten seconds of coning motion. The number of gyro samples is set to $N = d = 8$ with $\alpha = 1$ degree and $f_c = 50$ Hz. Both EFH rational and polynomial-BAC interpolations lead to smaller attitude errors by about two orders in the middle part of the interval. Specifically, there exists accuracy degradation in the first and the last computing interval of eight samples in which the normal polynomial interpolation has to be used for the sake of being at boundary intervals. The subfigure on the right hand presents the angular error if the accuracy assessment is performed by excluding the two boundary intervals. It is interesting to see that the polynomial-BAC interpolation outperforms the EFH rational interpolation even by about four orders, while the latter, as a blend of polynomials, has a similar error shape with the normal polynomial. It should be stressed that excluding the boundary intervals for accuracy assessment is not over-optimistic for those applications that the motion is benign at the start and in the end. Figure 7 plots the attitude error trend as a function of the coning frequency ratio relative to the sampling rate, namely $f_c/f_s$. The coning frequency ranges from 1 Hz to 200 Hz. The EFH rational and polynomial-BAC interpolations do not bring accuracy benefit relative to the polynomial interpolation until the coning frequency is larger than 50 Hz (Fig. 7, left subfigure). In contrast, when the accuracy evaluation starts from the second computation interval, the advantage of the EFH rational and polynomial-BAC interpolations is prominent, and notably the polynomial-BAC interpolation gaining about five orders of magnitude in accuracy as compared with the polynomial interpolation (Fig. 7, right subfigure). Specifically, the attitude error of the polynomial interpolation begins to quickly rise from 10 Hz onwards, while that of the polynomial-BAC interpolation does not significantly increase until 100 Hz and after. This turning frequency could be further postponed by using larger $N$, the number of gyro samples in one computation interval. As shown in the right subfigure of Fig. 7, $N = d = 16$ reduces the attitude error at 200 Hz by nearly six orders of magnitude relative to $N = d = 8$. In other words, the EFH rational and polynomial-BAC interpolations, especially the latter, help enhance the capability in coping with attitude computation of highly-dynamic motions.

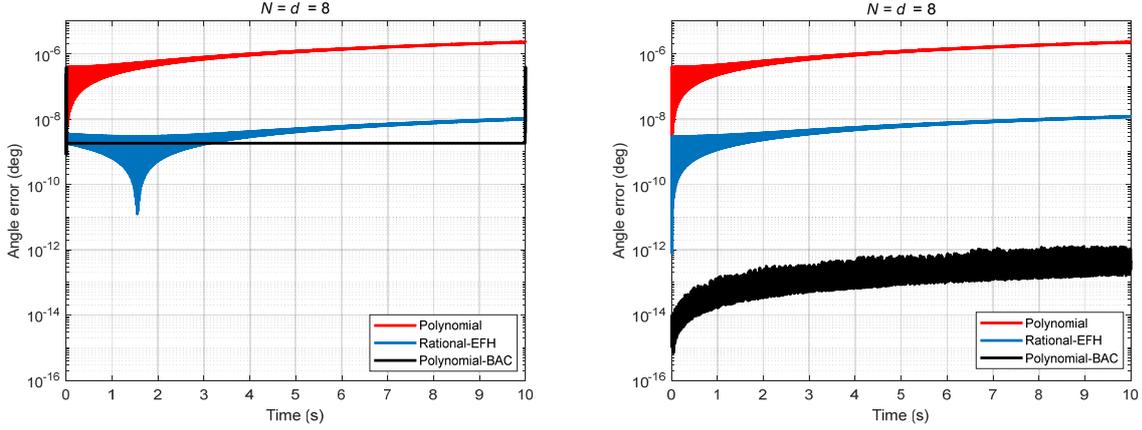

Figure 6. Attitude errors with $N = d = 8$ for polynomial, EFH rational and polynomial-BAC angular velocity reconstruction ($\alpha = 1$ deg, $f_c = 50$ Hz). The right-hand subfigure highlights the attitude errors when the two boundary intervals of $d$ samples are excluded from accuracy assessment.

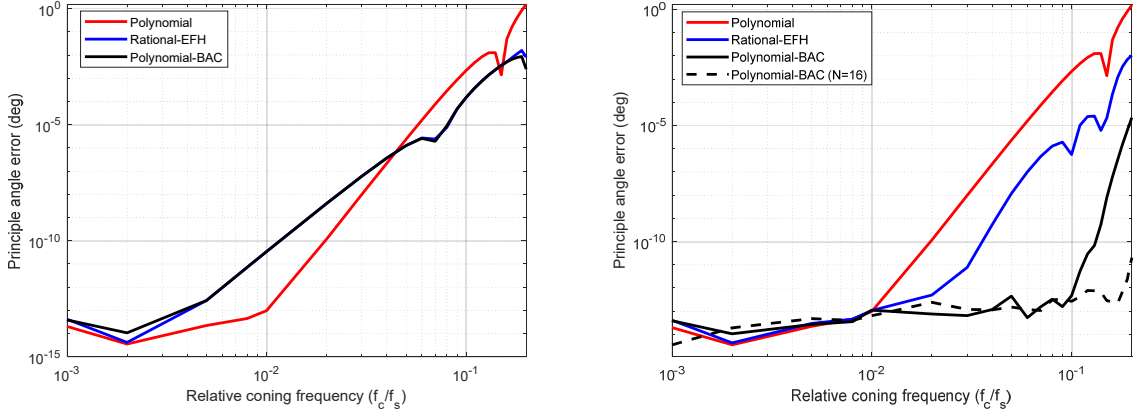

Figure 7. Attitude errors as function of relative frequency for polynomial, EFH rational and polynomial-BAC angular velocity reconstruction ($\alpha = 1$ deg. Solid lines for $N = d = 8$; dashed line for $N = d = 16$). The right-hand subfigure highlights the attitude errors when the two boundary intervals of $d$ samples are excluded from accuracy assessment.

Finally, practical situations with noisy gyroscope measurement are next investigated, as the high-order/sample algorithms tend to be much more sensitive to narrow-band noises that might lead to pseudo-coning [27]. Noise errors with angle random walks of 0.001 $\deg/\sqrt{h}$ and 0.2 $\deg/\sqrt{h}$ are considered, comparable to navigation-grade and tactical-grade inertial navigation systems, respectively. The coning angles are set to $\alpha = 1$ and 10 degrees, respectively. The first case simulates high vibration environment and the second case mimics highly dynamic motions like those experienced by smart munitions [28]. A common set of random gyroscope noises are generated and fed to all interpolation methods for a fair comparison. Figure 8 plots the results for error-contaminated measurements while excluding the boundary intervals in accuracy assessment. The considered coning frequency ranges from 10 Hz to 200 Hz. The error behaviors are quite the same in both cases. For coning frequencies below 60 Hz, the gyroscope measurement errors dominate the attitude computation accuracy and the attitude errors are flat lines for all interpolations. The normal and EFH rational interpolations begin to rise at 60 Hz and 150 Hz, respectively, where the computation errors start to take the lead, while the polynomial-BAC interpolation just keeps flat in the whole range of considered coning frequencies. Compared with the normal polynomial interpolation at the coning frequency of 200 Hz, the EFH rational interpolation reduces the attitude errors by about two orders, and the polynomial-BAC interpolation reduces by about four orders, totally insensitive to the relative coning frequency in the considered range. It demonstrates the powerfulness of the BAC-based interpolations in depressing the Runge effect in attitude computation.

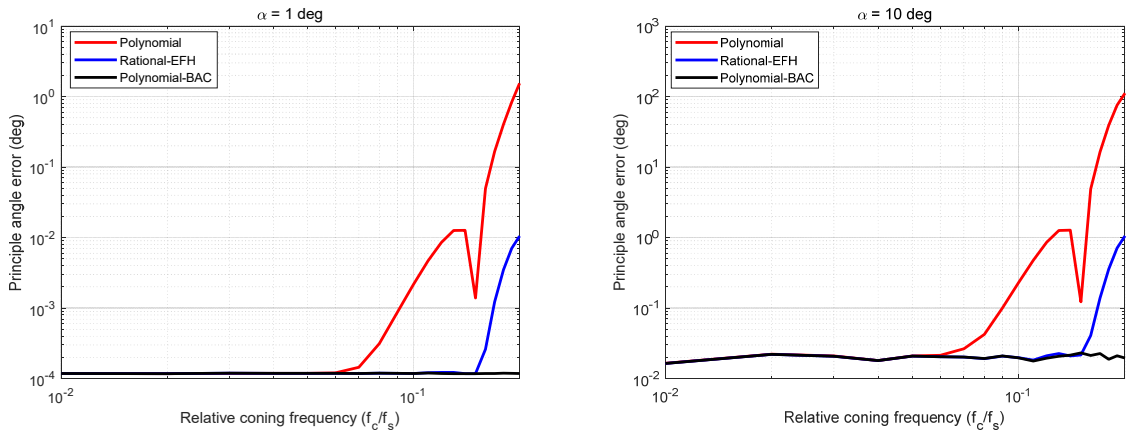

Figure 8. Attitude errors as function of relative frequency for polynomial, EFH rational and polynomial-BAC angular velocity reconstruction ( $N = d = 8$ ) for noise-contaminated gyro measurements (left: 0.001 $\deg/\sqrt{h}$ , right: 0.2 $\deg/\sqrt{h}$ ). The two boundary intervals of $d$ samples are excluded from accuracy assessment.

## IV. CONCLUSIONS

The Runge effect is a frequently-encountered problem in interpolations on equispaced samples. This work reviews its underlying rationale, as well as the polynomial and rational interpolants in the barycentric form. The BAC strategy using additional measurements in the neighbouring intervals is shown effective to reduce the Lebesgue constant of the rational interpolant. Inspired by this observation, we investigate the BAC strategy in polynomial interpolation and find that the resultant polynomial-BAC interpolation is even more effective in depressing the Runge effect. Both the EFH rational and polynomial-BAC interpolations are applied to reconstruct the angular velocity from gyroscope angular incremental measurements for attitude computation using the recent functional iteration approach. Results under the classical coning motion show that the polynomial-BAC interpolation outperforms the EFH rational interpolation in the context of attitude computation, and can significantly enhance the capability in coping with attitude computation for sustained highly-dynamic motions. It would also potentially bring significant benefits to the inertial navigation computation.

## ACKNOWLEDGEMENTS

Thanks to PhD student Hongyan Jiang, at Shanghai Key Laboratory of Navigation and Location-based Services, for reading an early draft.

## REFERENCES

[1] F. L. Markley, J. L. Crassidis, *Fundamentals of Spacecraft Attitude Determination and Control*: Springer, 2014.

[2] P. D. Groves, *Principles of GNSS, Inertial, and Multisensor Integrated Navigation Systems*, 2nd ed.: Artech House, Boston and London, 2013.

[3] "A study of the critical computational problems associated with strapdown inertial navigation systems," NASA CR-968 by United Aircraft Corporation, 1968.

[4] R. Miller, "A new strapdown attitude algorithm," *Journal of Guidance, Control, and Dynamics,* vol. 6, pp. 287-291, 1983.

[5] V. Z. Gusinsky, V. M. Lesyuchevsky, Y. A. Litmanovich, H. Musoff, G. T. Schmidt, "New procedure for deriving optimized strapdown attitude algorithm," *Journal of Guidance, Control, and Dynamics,* vol. 20, pp. 673-680, 1997.

[6] Y. A. Litmanovich, V. M. Lesyuchevsky, V. Z. Gusinsky,

"Two new classes of strapdown navigation algorithms," *Journal of Guidance, Control, and Dynamics,* vol. 23, pp. 34-44, 28-30, Jun. 2000.

[7] P. G. Savage, "A unified mathematical framework for strapdown algorithm design," *Journal of Guidance Control and Dynamics,* vol. 29, pp. 237-249, 2006.

[8] Y. Wu, "RodFIter: attitude reconstruction from inertial measurement by functional iteration," *IEEE Trans. on Aerospace and Electronic Systems,* vol. 54, pp. 2131-2142, 2018.

[9] Y. Wu, Q. Cai, T.-K. Truong, "Fast RodFIter for attitude reconstruction from inertial measurement," *IEEE Trans. on Aerospace and Electronic Systems,* vol. 55, pp. 419-428, 2019.

[10] Y. Wu, G. Yan, "Attitude reconstruction from inertial measurements: QuatFIter and its comparison with RodFIter," *IEEE Trans. on Aerospace and Electronic Systems,* vol. 55, pp. 3629-3639, 2019.

[11] Y. Wu, "iNavFIter: Next-generation inertial navigation computation based on functional iteration," *IEEE Trans. on Aerospace and Electronic Systems,* vol. 56, pp. 2061-2082, 2020.

[12] J.-P. Berrut, G. Klein, "Recent advances in linear barycentric rational interpolation," *Journal of Computational and Applied Mathematics,* vol. 259, pp. 95–107, 2014.

[13] L. N. Trefethen, *Approximation Theory and Approximation Practice*: SIAM, 2012.

[14] R. B. Platte, L. N. Trefethen, A. B. J. Kuijlaars, "Impossibility of Fast Stable Approximation of Analytic Functions from Equispaced Samples," *SIAM Review,* vol. 53, pp. 308-318, 2011.

[15] M. S. Floater, K. Hormann, "Barycentric rational interpolation with no poles and high rates of approximation," *Numerische Mathematik,* pp. 315-331, 2007.

[16] G. Klein, "An extension of the Floater–Hormann family of barycentric rational interpolants," *Mathematics of Computation,* vol. 82, pp. 2273-2292, 2013.

[17] M. Ignagni, "Enhanced Strapdown Attitude Computation," *Journal of Guidance Control and Dynamics,* vol. 43, pp. 1220–1224, 2020.

[18] M. Ignagni, "Accuracy Limits on Strapdown Velocity and Position Computations," *Journal of Guidance Control and Dynamics,* vol. 44, pp. 654-658, 2020.

[19] Y. Wu, Y. A. Litmanovich, "Strapdown attitude computation: functional iterative integration versus taylor series expansion," *Gyroscopy and Navigation,* vol. 11, pp. 263-276, 2020.

[20] W. H. Press, *Numerical Recipes: the Art of Scientific Computing,* 3rd ed. Cambridge; New York: Cambridge University Press, 2007.

[21] S. Guttel, G. Klein, "Convergence of linear barycentric rational interpolation for analytic functions," *SIAM Journal on Numerical Analysis,* vol. 50, pp. 2560-2580, 2012.

[22] L. Bos, S. D. Marchi, K. Hormann, G. Klein, "On the Lebesgue constant of barycentric rational interpolation at equidistant nodes," *Numerische Mathematik,* vol. 121, pp. 461-471, 2012.

[23] Y. F. Jiang, Y. P. Lin, "Improved strapdown coning algorithms," *IEEE Trans. on Aerospace and Electronic Systems,* vol. 28, pp. 484-490, 1992.

[24] M. B. Ignagni, "Efficient class of optimized coning compensation algorithm," *Journal of Guidance, Control, and Dynamics,* vol. 19, pp. 424-429, 1996.

[25] P. G. Savage, "Strapdown inertial navigation integration algorithm design, part 1: attitude algorithms," *Journal of Guidance, Control, and Dynamics,* vol. 21, pp. 19-28, 1998.

[26] T. J. Rivlin, *The Chebyshev Polynomials*: John Wiley & Sons Inc., 1974.

[27] Y. A. Litmanovich, "Use of angular rate multiple integrals as input signals for strapdown attitude algorithms," in *Symposium Gyro Technology*, Stuttgart, Germany, 1997.

[28] R. Holm, H. R. Petersen, S. Normann, H. Schou, M. Horntvedt, M. Hage, S. Martinsen, "High–g (20,000g+) testing of an existing tactical grade gyro design," in *DGON Inertial Sensors and Systems (ISS)*, Braunschweig, Germany, 2020.